\documentclass{article}

    \usepackage[final]{neurips_2025}

\usepackage[utf8]{inputenc} 
\usepackage[T1]{fontenc}    
\usepackage{hyperref}       
\usepackage{url}            
\usepackage{booktabs}       
\usepackage{amsfonts}       
\usepackage{nicefrac}       
\usepackage{microtype}      
\usepackage{xcolor}         
\usepackage{graphicx}
\usepackage{algorithm2e}
\RestyleAlgo{ruled}
\usepackage{amsmath} 
\usepackage{pifont}
\usepackage{wrapfig}
\usepackage{subcaption} 
\usepackage{multirow} 
\usepackage{array}    

\title{Subject or Style: Adaptive and Training-Free Mixture of LoRAs}

\author{%
  Jia-Chen Zhang \\
  School of Electronic and Electrical Engineering\\
  Shanghai University of Engineering Science\\
  \texttt{m325123603@sues.edu.cn} \\
  \And
  Yu-Jie Xiong \\
  School of Electronic and Electrical Engineering\\
  Shanghai University of Engineering Science\\
  \texttt{xiong@sues.edu.cn}\\
}

\begin{document}

\maketitle

\begin{abstract}
Fine-tuning models via Low-Rank Adaptation (LoRA) demonstrates remarkable performance in subject-driven or style-driven generation tasks. Studies have explored combinations of different LoRAs to jointly generate learned styles and content. However, current methods struggle to balance the original subject and style, and often require additional training. Recently, K-LoRA proposed a training-free LoRA fusion method. But it involves multiple hyperparameters, making it difficult to adapt to all styles and subjects. In this paper, we propose EST-LoRA, a training-free adaptive LoRA fusion method. It comprehensively considers three critical factors: \underline{E}nergy of matrix, \underline{S}tyle discrepancy scores and \underline{T}ime steps. Analogous to the Mixture of Experts (MoE) architecture, the model adaptively selects between subject LoRA and style LoRA within each attention layer. This integrated selection mechanism ensures balanced contributions from both components during the generation process. Experimental results show that EST-LoRA outperforms state-of-the-art methods in both qualitative and quantitative evaluations and achieves faster generation speed compared to other efficient fusion approaches. Our code is publicly available at: \url{https://anonymous.4open.science/r/EST-LoRA-F318}.
\end{abstract}

\section{Introduction}
Diffusion models achieve high-quality sample generation by learning the distribution of target data and have demonstrated impressive performance in fields such as images, audio, and video \cite{NEURIPS2020_4c5bcfec,song2021denoising,Rombach_2022_CVPR}. With the emergence of various personalization methods, including StyleDrop \cite{10.5555/3666122.3669042} and DreamBooth \cite{Ruiz_2023_CVPR}, the flexibility of generative models has been significantly enhanced. Among these, LoRA enables personalized generation with only a few images of a specific subject or style \cite{hu2022lora}. It freezes the parameters of the base model and learns two low-rank incremental matrices $A$ and $B$. The plug-and-play feature and resource-efficient design of LoRA make it widely adopted.

Style transfer aims to combine the subject information of a content image with artistic features (e.g., Impressionist brushstrokes, ink wash rendering) from a style image to generate novel outputs that integrate both characteristics \cite{Gatys_2016_CVPR, Chung_2024_CVPR, Chen_2024_CVPR}. As LoRA-based personalization grows in popularity, researchers have begun exploring the fusion of LoRA weights \cite{Hartley_2024_CVPR, huang2024incontextloradiffusiontransformers}. However, directly applying LoRA to style transfer faces two critical challenges: 1) Generating a specific subject within a particular style often lacks sufficient training data, as obtaining images of a specific subject under a target style for personalized training is typically impractical. 2) Existing pre-trained LoRA weights for specific subjects and styles already exist, and retraining them would result in significant computational redundancy.

\begin{figure}[t]
  \centering
  \includegraphics[width=1\linewidth]{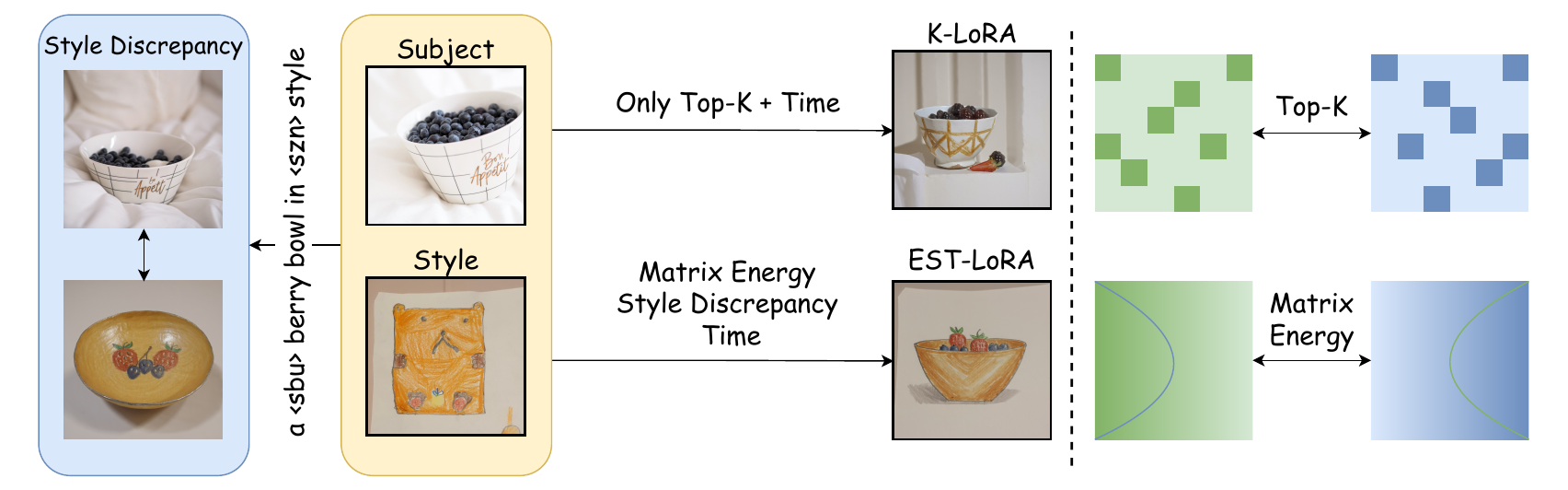}
  \caption{A simple demonstration of the difference between EST-LoRA and K-LoRA. EST-LoRA selects different LoRAs for inference based on Matrix Energy, Style Discrepancy Scores, and Time-Step, without requiring any training.}
  \label{photoup}
\end{figure}

Early studies employed variable coefficient grid search followed by subjective human evaluations to identify optimal combination strategies, but this approach is extremely time-consuming \cite{von-platen-etal-2022-diffusers}. Subsequently, ZipLoRA proposed a low-parameter method to determine suitable fusion ratios through minimal training \cite{10.1007/978-3-031-73232-4_24}. B-LoRA facilitates style transfer using only two attention modules \cite{10.1007/978-3-031-72684-2_11}. K-LoRA proposes a selection-based strategy that dynamically chooses between Subject LoRA and Style LoRA based on the Top-K of the target LoRA at different time steps during inference \cite{ouyang2025kloraunlockingtrainingfreefusion}, as shown in Figure \ref{photoup}. However, methods relying solely on time steps and hyperparameter selection struggle to perform consistently across all scenarios, resulting in poor generalizability.

Therefore, to address the issues of high resource consumption and poor generalization ability in existing models, this paper proposes a training-free style transfer method based on LoRA fusion, dubbed EST-LoRA. It jointly selects LoRA weights based on matrix energy, style discrepancy scores, and time steps. Notably, EST-LoRA requires only a single hyperparameter to achieve training-free arbitrary LoRA fusion, whereas K-LoRA requires two hyperparameters, one of which needs to be adjusted for different LoRAs. Our contributions are summarized as follows:
\begin{itemize}
\item[$\bullet$]We propose EST-LoRA, a simple and novel style transfer method that achieves training-free performance by adaptively selecting the appropriate target LoRA. Compared to other training-free baseline methods, it achieves a 5\% improvement in performance and a 30\% increase in speed.
\end{itemize}
\begin{itemize}
\item[$\bullet$]We comprehensively integrate three key factors influencing style transfer quality: style discrepancy scores, matrix energy, and time steps. This enables adaptive generalization across arbitrary styles or subjects.
\end{itemize}
\begin{itemize}
\item[$\bullet$]Extensive experiments demonstrate that EST-LoRA outperforms existing baselines in both performance and speed.
\end{itemize}

\section{Related work}
Image-based style transfer is a research field that dates back at least two decades \cite{10.1145/383259.383296, 10.1145/383259.383295}. CNN-based methods have made significant progress in arbitrary style transfer \cite{Gatys_2016_CVPR, 10.1007/978-3-319-46475-6_43, NIPS2017_49182f81}. Generative models such as GANs have also been employed as prior models for image stylization tasks \cite{Karras_2019_CVPR, 10.1145/3528223.3530164, Ojha_2021_CVPR, zhu2022mind}. Recently, many GAN-based methods have successfully achieved one-shot stylization by fine-tuning pretrained GANs on given reference styles.

Since the emergence of diffusion models, the field of Text-to-Image (T2I) generation has undergone remarkable advancements. Diffusion models, with their robust generative capabilities and capacity to model complex data distributions, have rapidly become a mainstream approach for image generation tasks, demonstrating exceptional performance \cite{NEURIPS2020_4c5bcfec,song2021denoising,Rombach_2022_CVPR}. However, challenges persist in generating images with specific subjects or styles. Recent studies have introduced various methods to fine-tune large-scale T2I diffusion models. Works such as \cite{gal2023an, Ruiz_2023_CVPR, Kumari_2023_CVPR} adapt models to new tasks by adjusting their original parameters. DreamArt \cite{Kumari_2023_CVPR} achieves self-supervised training through joint positive and negative embeddings, while LoRA \cite{hu2022lora} and StyleDrop \cite{10.5555/3666122.3669042} enable style personalization via low-rank incremental matrices and a small subset of weights, respectively. Among these, LoRA is widely adopted due to its plug-and-play nature and full freezing of the base model parameters.

As an increasing number of stylistically diverse LoRA weights become publicly available, researchers have begun exploring combinations of different LoRA modules to achieve style fusion and transfer. Early methods focused on training new LoRA modules from scratch based on original LoRA configurations, but this approach remains computationally expensive for users. Recent efforts have shifted toward resource-efficient fusion strategies. For instance, \cite{wu2024mixture} employs gating functions to combine multiple LoRAs, while ZipLoRA freezes all LoRA weights and the base model, training only a minimal set of parameters to compute weighted sums of original LoRAs \cite{10.1007/978-3-031-73232-4_24}. B-LoRA \cite{10.1007/978-3-031-72684-2_11} decouples subject and style representations by training two core attention modules within the LoRA framework. K-LoRA compares LoRA matrix importance through Top-K parameter selection and controls significance scores via hyperparameters and time steps, achieving training-free fusion \cite{ouyang2025kloraunlockingtrainingfreefusion}. However, these methods exhibit limited generalization across style disparities due to reliance on fixed hyperparameters and manual tuning. In contrast, our method comprehensively considers style discrepancy scores, matrix energy, and time steps to enable adaptive, training-free style transfer for arbitrary style differences.

\section{Method}
\label{method}
\subsection{Diffusion model}
Diffusion Model represents a significant breakthrough in the field of generative models in recent years. It models data distributions through a gradual denoising mechanism, with its core idea originating from the diffusion process in non-equilibrium thermodynamics. The essence of the Diffusion Model comprises two key components: the forward noising process and the reverse generative process. The forward diffusion process is defined by a Markov chain, where the addition of noise at each step can be expressed as:
\begin{equation}
\label{eq:1}
q(\mathbf{x}_t \mid \mathbf{x}_{t-1}) = \mathcal{N} \left( \mathbf{x}_t; \sqrt{1-\beta_t}\, \mathbf{x}_{t-1}, \beta_t \mathbf{I} \right),
\end{equation}
where $\beta_t$ is the predefined noise variance, and $\mathbf{x}_t$ represents the noisy data at step $t$. After $T$ steps of diffusion, the data approaches a standard Gaussian distribution. The reverse generative process, on the other hand, uses a neural network to predict the noise or data, progressively reconstructing the sample.

\subsection{Matrix energy}
\begin{wrapfigure}{r}{0.5\linewidth}
    \centering
    \includegraphics[width=\linewidth]{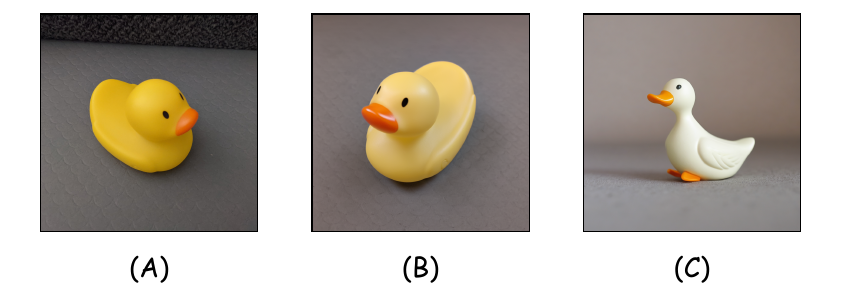}
    \caption{Illustration of the impact of different matrix pruning strategies on the generation results. (a) The original generation result serves as the control. (b) The generation result after pruning the largest values in the matrix. (c) The generation result after pruning the largest singular values.}
    \label{photo}
\end{wrapfigure}
In the field of deep learning model parameter optimization, the distribution characteristics of matrix energy significantly influence model performance. Previous studies have shown that selecting the Top-K elements from a matrix for feature extraction can effectively preserve model performance \cite{10.1007/978-3-031-73232-4_24, ouyang2025kloraunlockingtrainingfreefusion}. However, this strategy exhibits notable limitations in visual generation tasks such as style transfer. Ablation experiments presented in Figure \ref{photo} demonstrate that retaining only the largest singular value preserves the basic image structure but results in the loss of fine texture details. In contrast, completely removing the largest singular value leads to significant distortion of the overall style. This observation supports the key finding of study \cite{zhang2023adaptive,NEURIPS2024_db36f4d6}: the largest singular value dominates global semantic representation, while the set of secondary singular values collectively determines the quality of local texture features.

It is worth noting that traditional matrix energy evaluation methods based on singular value decomposition (SVD) face significant computational bottlenecks. The time complexity of SVD operations is O($n^{3}$), which often exceeds the practical computational budget in high-resolution image processing scenarios. To address this efficiency challenge, the EST-LoRA framework proposed in this study introduces the Frobenius norm as an alternative metric. The Frobenius norm is defined as the Euclidean norm of all matrix elements, and its mathematical expression is given by:
\begin{equation}
||A||_F = \sqrt{\sum_{i=1}^m \sum_{j=1}^n |a_{ij}|^2},
\end{equation}
This norm offers three key advantages: (1) its computational complexity is only O(mn), representing a reduction by an order of magnitude compared to SVD; (2) mathematically, it equals the sum of squares of all singular values: 
\begin{equation}
\|A\|_F^2 = \sum_{i=1}^{m} \sum_{j=1}^{n} |a_{ij}|^2 
= Tr(A^\top A)
= \sum_{i=1}^{r} \sigma_i^2,
\end{equation}
$Tr(\cdot)$ denotes the trace. $\sigma_i$ represents the $i$-th singular value of matrix A. Thus effectively reflecting the overall energy distribution of the matrix; (3) it does not require explicit computation of eigenvectors, thereby avoiding the additional cost associated with orthogonal basis transformation.

\begin{figure}[t]
  \centering
  \includegraphics[width=1\linewidth]{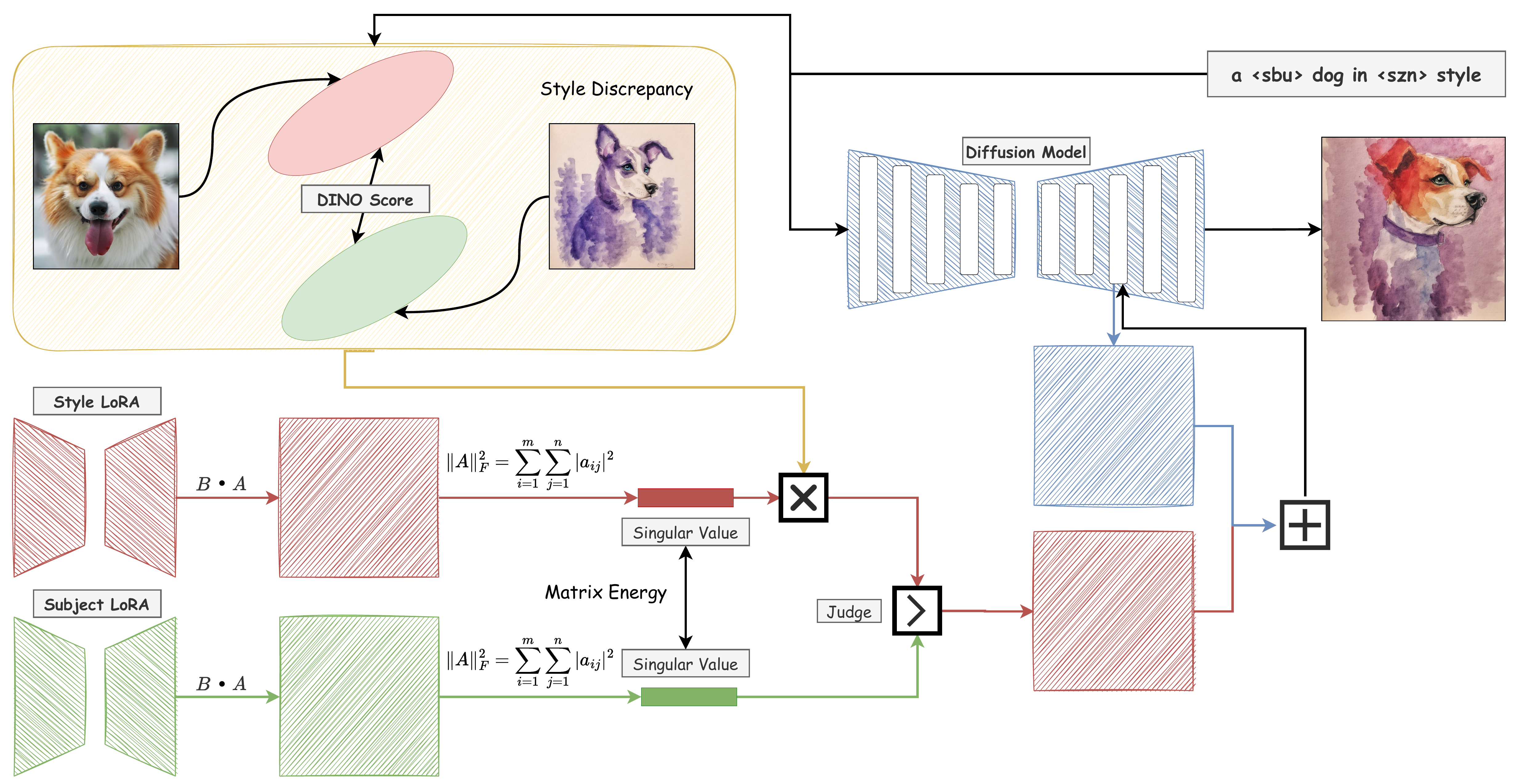}
  \caption{Overview of the proposed EST-LoRA. Before inference begins, EST-LoRA generates one image each using the target Subject LoRA and Style LoRA with the same prompt to evaluate the Style discrepancy. During inference, EST-LoRA selects the activated LoRA weights based on the Energy of matrix, Style discrepancy scores and Time steps.}
  \label{photoex}
\end{figure}

\subsection{EST-LoRA}
This section systematically outlines the core workflow of the EST-LoRA framework. As illustrated in Figure \ref{photoex}, our approach first employs identical prompts to drive two models separately configured with topic LoRA and style LoRA modules. These models generate two distinct images under a unified random seed specification. The architecture incorporates a pre-trained DINO-ViT16 model \cite{Caron_2021_ICCV}, which quantifies stylistic discrepancies between generated images through a dedicated Style discrepancy scoring mechanism:
\begin{equation}
\label{eq:4}
D_{(s,c)}=Discrepancy_{(Style,Content)}=\| \Phi_\text{DINO}(Style) - \Phi_\text{DINO}(Content)\|_F^2
\end{equation}
During the denoising phase of the diffusion process, our method leverages the Mixture-of-Experts (MoE) paradigm to adaptively select between subject LoRA and Style LoRA modules in each attention layer, rather than merging LoRA matrices which would compromise their original information. However, conventional MoE architectures incorporate a trainable gating function for expert allocation across layers. The critical challenge lies in transforming this mechanism into a training-free adaptive gate capable of dynamically accommodating arbitrary subject-content pairs and style configurations.
Based on recent studies \cite{Patashnik_2023_ICCV, voynov2024p, ouyang2025kloraunlockingtrainingfreefusion}, we find that during image generation, earlier time steps emphasize content reconstruction whereas subsequent stages gradually enhance texture and style attributes. To align with the identified patterns in our research, EST-LoRA introduces a Linear Adaptation Factor $\gamma$ into the LoRA module selection mechanism. This approach ensures that the model prioritizes content reconstruction during initial diffusion stages while progressively emphasizing texture and detail restoration in subsequent denoising phases:
\begin{equation}
\label{eq:5}
\gamma=\alpha \cdot \frac{time_{(now)}}{time_{(all)}} +(1- D_{(s,c)}),
\end{equation}
where $time_{(now)}$ denotes the current time step, and $time_{(all)}$ represents the total number of time steps required for the complete generation. $\alpha$ are hyperparameters. $(1- D_{(s,c)})$ acts as an adaptive prior. When the style discrepancy between the fused LoRAs is large, the value of $(1- D_{(s,c)})$ increases adaptively, causing the target style LoRA to be assigned a higher selection weight at an earlier stage. Conversely, when the styles of the fused LoRAs are similar, the value of $(1- D_{(s,c)})$ decreases, leading to the target style being selected later and less frequently. This mechanism endows EST-LoRA with strong scalability and robustness, enabling it to effectively handle LoRAs with varying degrees of stylistic differences. The final selection mechanism for the Subject LoRA and Style LoRA is as follows:
\begin{equation}
E_{S}= \|W_{S}\|_F^2,
\end{equation}
\begin{equation}
E_{C}=\|W_{C}\|_F^2,
\end{equation}
\begin{equation}
\label{eq:8}
C(E_C, E_s) = 
\begin{cases} 
\Delta W_c, & \text{if } S_c \geq \gamma \cdot S_s \\ 
\Delta W_s, & \text{otherwise}
\end{cases}
\end{equation}
where $W_c$ and  $W_s$ represent the Content and Style LoRA Weight. For a clearer understanding of our approach, we present the inference procedure of EST-LoRA in Algorithm \ref{alg:est_lora}.

\begin{algorithm}[t]
\caption{EST-LoRA Inference Procedure}
\label{alg:est_lora}
\KwIn{
$\mathcal{P}$: Shared prompt for dual LoRA modules;\\
$T$: Total diffusion time steps;\\
$\mathcal{L}_c, \mathcal{L}_s$: Content/Style LoRA modules;\\
$\alpha$: Time-step weight hyperparameter;\\
}
\KwOut{ $\mathbf{x}_0$: Generated image with fused LoRA attributes;
}
\texttt{// Compute style discrepancy}\\
Initialize noisy image $\mathbf{x}_T \sim \mathcal{N}(0,\mathbf{I})$;\\
Generate dual images$ \mathbf{x}_c$ and $\mathbf{x}_s$;\\
Compute style discrepancy via DINO-ViT16 \cite{Caron_2021_ICCV} as \ref{eq:4};\\
\texttt{// Start inference}\\
\For{timestep $t$ from $T$ to $1$}{
    Calculate adaptation factor $\gamma$ as \ref{eq:5};\\
    Evaluate LoRA energy metrics $E_{S}$ and $E_{C}$;\\
    Adaptive selection as \ref{eq:8};\\
    Update attention layers with selected $\Delta W$;\\
    Perform denoising step: $\mathbf{x}_{t-1} \leftarrow \text{DiffusionStep}(\mathbf{x}_t, \Delta W)$;\\
}

\textbf{Return} Final image $\mathbf{x}_0$.
\end{algorithm}

\section{Experiment}
\label{exp}
\subsection{Experimental settings}
\textbf{Datasets.} Following the previous experimental setup, we conducted tests by training local LoRA under the conditions of $r=8$, epochs$=50$, and learning rate$=5e-5$. For subject LoRA, we employed the Dreambooth \cite{Ruiz_2023_CVPR} dataset containing $4-5$ images per subject category. For style LoRA, we utilized the dataset provided by Styledrop \cite{10.5555/3666122.3669042}. In each style category, only a single image was used for training purposes.

\textbf{Experimental details.} We conducted experiments using the SDXL v1.0 base model. We selected nine locally trained subject LoRA modules and nine style LoRA modules for cross-combination testing, resulting in a total of 36 experimental groups. We compare EST-LoRA with four popular parameter-efficient tuning methods, including Direct arithmetic merging, Joint training, B-LoRA \cite{10.1007/978-3-031-72684-2_11}, ZipLoRA \cite{10.1007/978-3-031-73232-4_24}, and K-LoRA \cite{ouyang2025kloraunlockingtrainingfreefusion}. We used the same hyperparameters as  \cite{ouyang2025kloraunlockingtrainingfreefusion}. For previous training-based methods, we only present the average of all results. In contrast, for K-LoRA, which is also training-free, we provide more detailed results. We use CLIP \cite{pmlr-v139-radford21a} to measure style similarity. Subject similarity is computed using both the CLIP score and DINO score \cite{zhang2023dino}.
\begin{table}
    \centering
    \tabcolsep=0.4cm
    \caption{Comparison of Methods and Results}
    \label{tab:results}
    \begin{tabular}{l c c c c }
        \toprule
        \textbf{Method} & \textbf{Style Sim} & \textbf{Clip Score} & \textbf{CLIP $\pm$} & \textbf{DINO Score}  \\
        \midrule
        \textbf{Direct} & 48.90\% & 66.60\% & 17.70\% & 43.00\% \\
        \textbf{Joint} & 68.20\% & 57.50\% & 10.70\% & 17.40\%  \\
        \textbf{B-LoRA} & 58.00\% & 63.80\% & 5.80\% & 30.60\%  \\
        \textbf{ZipLoRA} & 60.40\% & 64.40\% & 4.00\% & 35.70\%  \\
        \midrule
        \textbf{K-LoRA} & & & &  \\
        Berry Bowl & 57.61\% & 70.03\% &/ & 19.46\%  \\
        Can & 54.48\% & 61.04\% & / & 24.41\% \\
        Clock & 60.40\% & 79.32\% &/ & 21.59\%  \\
        Dog & 64.22\% & 75.96\% &/ & 35.50\% \\
        Monster Toy & 57.91\% & 71.35\% &/ & 31.08\%  \\
        Poop Emoji & 54.96\% & 78.32\% &/ & 25.95\%  \\
        Rc Car & 57.69\% & 76.38\% &/ & 24.67\%  \\
        Red Cartoon & 58.09\% & 70.49\% & / & 32.28\%  \\
        Cat & 66.37\% & 78.82\% &/ & 33.00\%  \\
        \midrule
        Average & 59.10\% & 73.50\% & 14.40\% & 27.50\%  \\
        \midrule
        \textbf{Ours} & & & &  \\
        Berry Bowl & 57.80\% & 69.21\% &/ & 23.41\%  \\
        Can & 61.90\% & 53.75\% &/ & 34.25\% \\
        Clock & 64.34\% & 73.82\% &/ & 31.11\%  \\
        Dog & 65.93\% & 71.63\% &/ & 38.02\%  \\
        Monster Toy & 58.74\% & 71.38\% &/ & 29.88\%  \\
        Poop Emoji & 56.62\% & 75.14\% &/ & 28.29\%  \\
        Rc Car & 62.34\% & 69.57\% &/ & 29.53\%  \\
        Red Cartoon & 66.19\% & 64.22\% &/ & 45.76\%  \\
        Cat & 64.77\% & 81.27\% &/ & 32.22\%  \\
        \midrule
        Average & 62.07\% & 70.00\% & 7.93\% & 32.50\% \\
        \bottomrule
    \end{tabular}
\end{table}

\subsection{Results}
In this section, we evaluate the effectiveness of EST-LoRA from both qualitative and quantitative perspectives.
\subsubsection{Quantitative comparisons}
The quantitative analysis of EST-LoRA demonstrates its competitive performance across multiple evaluation metrics, as shown in Table~\ref{tab:results}. This approach particularly exhibits superior capability in balancing style preservation, semantic alignment, and visual feature fidelity compared to baseline methods such as K-LoRA. Specifically, EST-LoRA achieves an average Style Sim score of 62.07\%, surpassing K-LoRA 59.10\% and indicating enhanced retention of target stylistic characteristics, with notable results observed in categories like Cat and Clock. While Ours exhibits a marginally lower Clip Score 70.00\% relative to K-LoRA 73.50\%, it demonstrates superior consistency in text-image alignment stability, as evidenced by its significantly reduced CLIP Score Difference (7.93\% vs. 14.40\%), which reflects diminished variability in semantic coherence across generated outputs. Furthermore, EST-LoRA outperforms K-LoRA in DINO Score (32.50\% vs. 27.50\%), highlighting its efficacy in capturing fine-grained visual features, particularly for challenging categories such as Red Cartoon and Can.

\subsubsection{Qualitative comparisons}
\begin{figure}[t]
  \centering
  \includegraphics[width=1\linewidth]{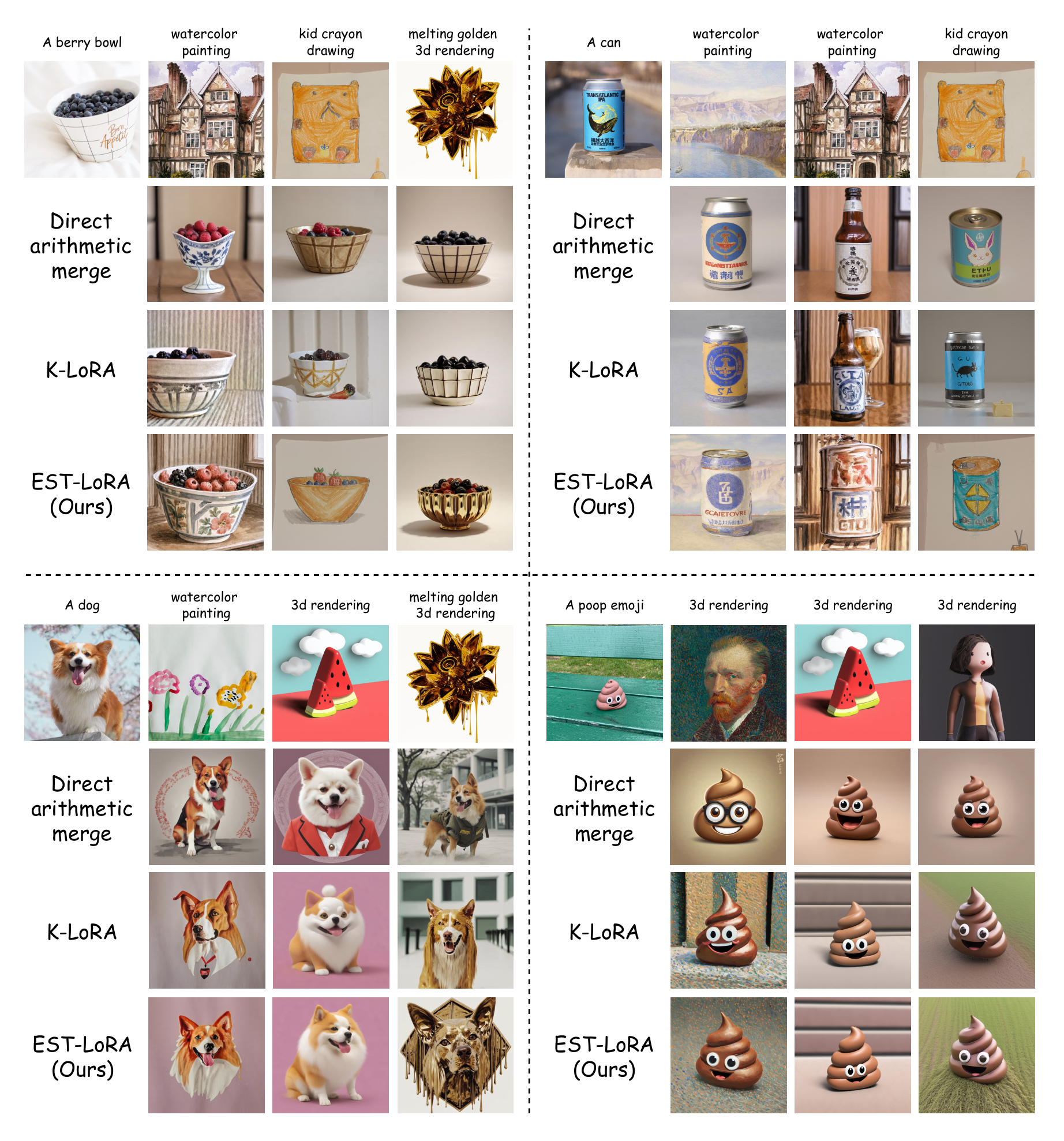}
  \caption{Qualitative Comparisons. We generate images using the same random seed. EST-LoRA balances subject fidelity and style adaptation by perceiving the style discrepancy.}
  \label{photomain}
\end{figure}

In this section, we evaluate the effectiveness of EST-LoRA based on the generated results. To ensure a fair comparison, we only assess training-free methods. All experiments use SDXL v1.0 as the base model, and results in the same column are generated using the same random seed. The results are shown in Figure \ref{photomain}.

The results indicate that Direct arithmetic merge struggles to control the subject of the generated output. For example, on datasets like "Berry Bowl" and "can", it fails to preserve the original subject characteristics. In contrast, K-LoRA significantly improves subject control and retains the original features across nearly all results. However, due to its limited sensitivity to style, K-LoRA relies on fixed hyperparameters during inference and therefore struggles with large style discrepancies. For instance, it performs poorly in tasks involving styles such as "melting golden 3D rendering" or "kid crayon drawing". EST-LoRA effectively balances subject fidelity and style adaptation. It dynamically adjusts the contribution of the style LoRA during inference based on the style discrepancy, achieving the best overall performance across different scenarios.

\begin{table}
    \centering
    \tabcolsep=0.4cm
    \caption{Results of the efficiency analysis. The experiment uses SDXL to generate an image of 1024 resolution.}
    \label{tab:time}
    \begin{tabular}{l|cccc}
        \toprule
        \textbf{Method}  & \textbf{Only Style}&\textbf{Direct}&\textbf{K-LoRA}&\textbf{EST-LoRA}\\
        \midrule
        \textbf{Time (s/photo)}& 15&25&34&26\\
        \bottomrule
    \end{tabular}
\end{table}

\begin{table}
    \centering
    \caption{Ablation Experiment Results. We conduct ablation experiments on the Energy of matrix, Style discrepancy scores, and Time steps.}
    \label{tab:abl}
    \begin{tabular}{l|ccc|c c c c }
        \toprule
        \textbf{Method}  & \textbf{Energy}& \textbf{Style}& \textbf{Time}& \textbf{Style Sim} & \textbf{Clip Score} &  \textbf{DINO Score}  \\
        \midrule
        \textbf{Direct}& \ding{55}& \ding{55}& \ding{55} & 48.90\% & 66.60\% &22.40\% \\
        \textbf{K-LoRA}& \ding{55}& \ding{55}& \ding{51} & 72.17\% & 69.30\% &36.98\% \\
        \textbf{Ours-ET} &\ding{51} & \ding{55} & \ding{51} &74.44\% &67.30\% &40.36\%\\
        \textbf{Ours-ST} &\ding{55} & \ding{51}  & \ding{51} &72.43\% &69.04\% &37.13\% \\
        \textbf{Ours-EST} & \ding{51} & \ding{51} & \ding{51} & \textbf{74.21\%} &\textbf{69.61\%} &\textbf{41.12\%} \\
        \bottomrule
    \end{tabular}
\end{table}

\subsection{Efficiency analysis}
In this section, we analyze the efficiency of EST-LoRA. The results are shown in Table~\ref{tab:time}. To ensure a fair comparison, all methods use the same set of hyperparameters. Only Style achieves the fastest processing speed (15 s/photo), as it does not involve any LoRA merging and only uses a single LoRA for generation. Direct arithmetic merge combines two LoRAs using fixed weights, resulting in increased computational cost (25 s/photo). In comparison, both K-LoRA and EST-LoRA introduce more sophisticated weight selection mechanisms to improve model performance. Among them, K-LoRA selects Top-K most suitable LoRA modules based on their matrix values, which results in the highest inference delay (34 s/photo). EST-LoRA chooses the optimal LoRA combination based on the Frobenius norm of the LoRA matrices. It achieves performance comparable to K-LoRA while maintaining a relatively high level of efficiency (26 s/photo). This indicates that EST-LoRA strikes a better balance between performance and computational efficiency.

\subsection{Ablation analysis}
\begin{wrapfigure}{r}{0.5\linewidth}
  \centering
  \includegraphics[width=1\linewidth]{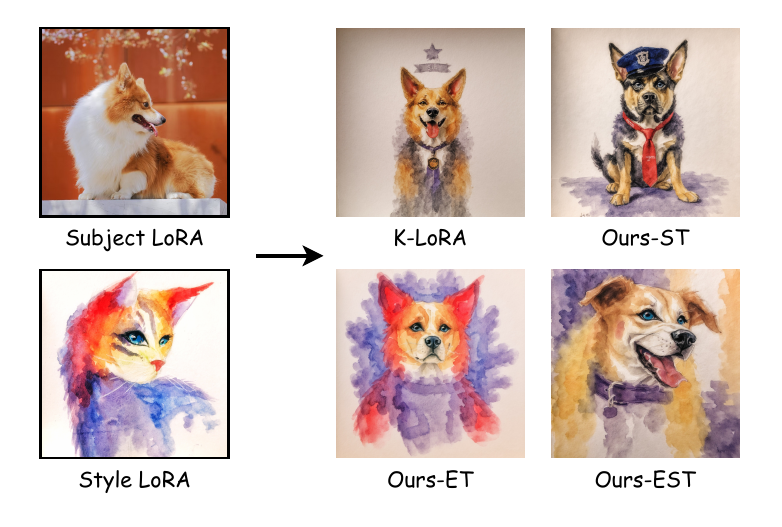}
  \caption{Qualitative Comparisons. We generate images using the same random seed. EST-LoRA balances subject fidelity and style adaptation by perceiving the style discrepancy.}
  \label{photoabl}
\end{wrapfigure}
In this section, we conduct a comprehensive ablation study to systematically evaluate the contribution of each component in our framework. All experiments are performed under the same set of hyperparameters. The results are shown in Table \ref{tab:abl}, where the method that jointly considers Energy of matrix, Style discrepancy scores and Time steps achieves the best overall performance. From the analysis, we find that using singular values to measure the Energy of matrix tends to focus more on the style details of the generated images, whereas using Top-K emphasizes the subject more. The results in Figure~\ref{photoabl} provide evidence for this argument. 

\begin{figure}[t]
  \centering
  \begin{subfigure}{\linewidth}
    \centering
    \includegraphics[width=1\linewidth]{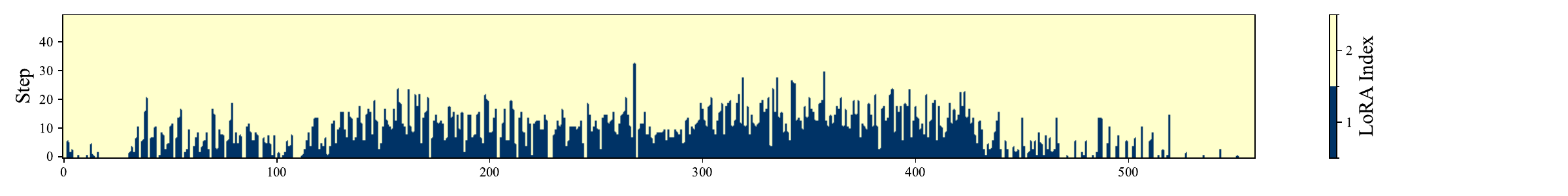}
    \caption{}
    \label{fig:A}
  \end{subfigure}

  \begin{subfigure}{\linewidth}
    \centering
    \includegraphics[width=1\linewidth]{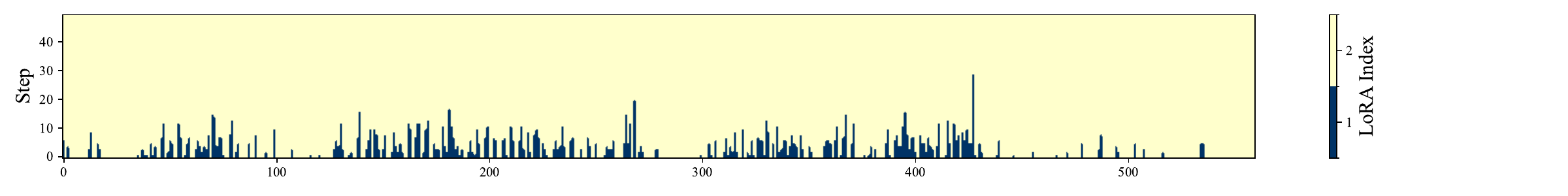}
    \caption{}
    \label{fig:B}
    
  \end{subfigure}
    \begin{subfigure}{\linewidth}
    \centering
    \includegraphics[width=1\linewidth]{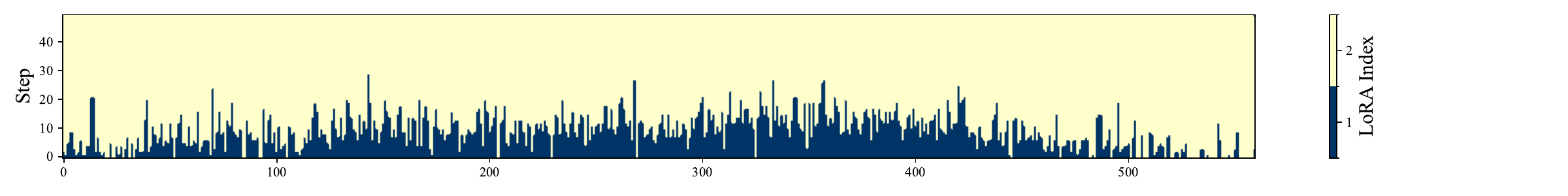}
    \caption{}
    \label{fig:C}
    
  \end{subfigure}
    \begin{subfigure}{\linewidth}
    \centering
    \includegraphics[width=1\linewidth]{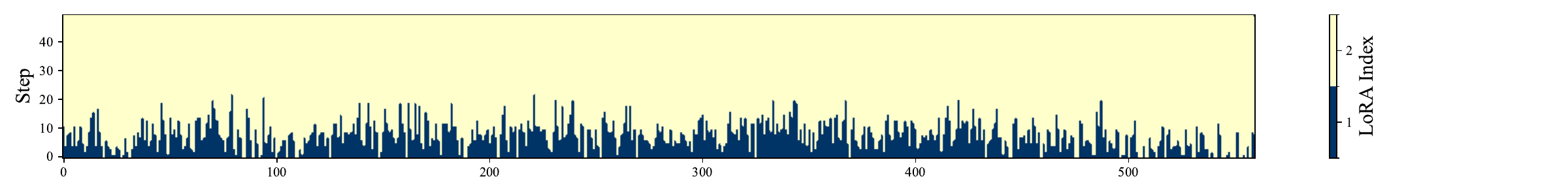}
    \caption{}
    \label{fig:D}
  \end{subfigure}
  \caption{Visualization of the layers selected by EST-LoRA during inference. This figure illustrates the selection results for each layer at each time step. The vertical axis represents the total of 50 diffusion steps, and the horizontal axis represents the total of 560 layers.  LoRA Index 1 denotes the Style LoRA, and 2 denotes the Subject LoRA. (a) corresponds to the case with small style discrepancy ($D_{(s,c)}=0.40$). (b) corresponds to the case with large style discrepancy ($D_{(s,c)}=0.13$). (c) is the result of (a) on K-LoRA. (d) is the result of (b) on K-LoRA.}
\end{figure}

To better explain the selection process of the LoRA weights, we visualize the selection ratio during inference. The Subject LoRA and Style LoRA are seamlessly interleaved and mutually fused throughout the generation process. EST-LoRA adaptively controls the selection ratio of Style LoRA based on the style discrepancy scores. When the style discrepancy is small \ref{fig:A}, Style LoRA is introduced earlier into the generation process; whereas when the style discrepancy is large \ref{fig:B}, Style LoRA plays a more prominent role at later stages. This mechanism enables EST-LoRA to achieve better generalization. In contrast, K-LoRA has a weaker perception of different style discrepancies, and the selection ratio of Style LoRA remains similar, as shown in Figures \ref{fig:C} and \ref{fig:D}.

\section{Conclusion}
In this paper, we propose a novel training-free style transfer method named EST-LoRA. Our approach systematically incorporates three key factors during inference to maximize the utilization of the original LoRA weights. These factors include matrix energy based on singular values, style discrepancy scores derived from DINO distance, and diffusion time steps. By jointly leveraging these components to control the selection of different LoRA modules, the number of hyperparameters required in our training-free framework is reduced to just one. Moreover, by pre-evaluating the style discrepancy scores before inference, EST-LoRA achieves enhanced scalability and generalization. Extensive experiments demonstrate that EST-LoRA outperforms baseline methods both quantitatively and qualitatively. Compared to existing training-free approaches, our method achieves a 5\% improvement in performance (measured by DINO score) and a 30\% increase in generation speed.

\section{Limitations}
\label{limitation}
There are two limitations in this work. First, although this study has reduced the number of hyperparameters to just one, it still relies on hyperparameter tuning during inference, leaving room for further optimization. Second, despite achieving state-of-the-art performance among training-free methods, there remains a performance gap compared to other approaches that are based on training. We look forward to addressing these limitations in future research.

\bibliographystyle{unsrt}
\bibliography{our}

\clearpage
\newpage
\appendix

\section{Frobenius Norm}
The Frobenius norm is a fundamental and widely used norm in matrix analysis, known for its intuitive definition and desirable mathematical properties. For a complex matrix A of size m×n, the Frobenius norm is defined as the square root of the sum of the squares of the absolute values of all its elements:
\begin{equation}
||A||_F = \sqrt{\sum_{i=1}^m \sum_{j=1}^n |a_{ij}|^2},
\end{equation}
This norm can also be expressed in terms of the trace of a matrix as:
\begin{equation}
\|A\|_F = \sqrt{\text{Tr}(A^\dagger A)}
\end{equation}
where $ A^\dagger $ denotes the conjugate transpose of $ A $; furthermore, it is closely related to the singular values of the matrix --- if $ \sigma_1, \sigma_2, \dots, \sigma_r $ are the singular values of $ A $. So, the square root of the sum of the squares of all singular values:
\begin{equation}
\|A\|_F = \sqrt{\sum_{i=1}^r \sigma_i^2}
\end{equation}

\section{Datasets}
\subsection{Style datasets}
\begin{figure}[h]
  \centering
  \includegraphics[width=1\linewidth]{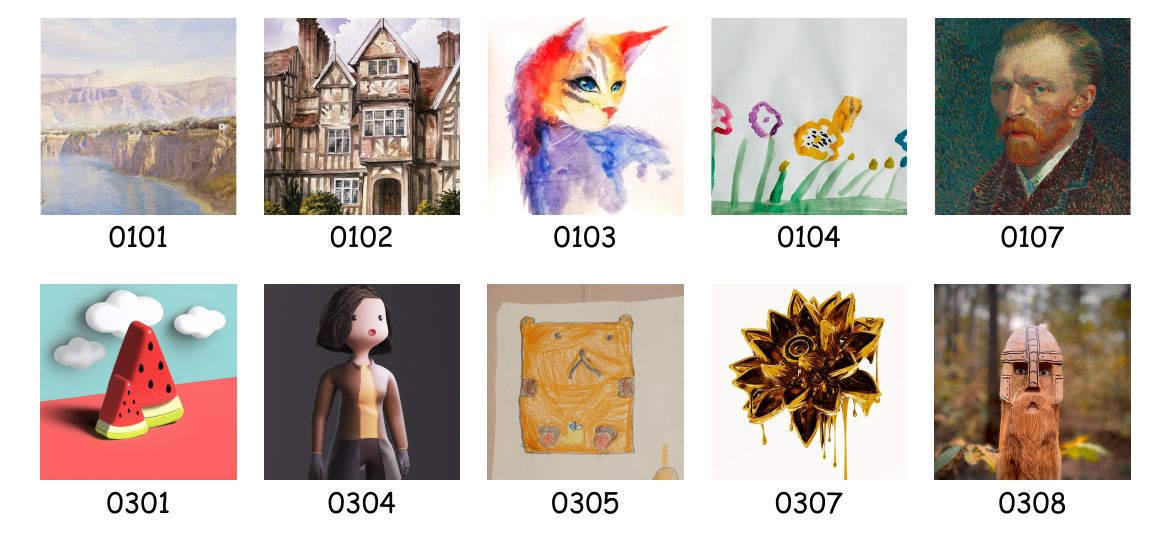}
  \caption{style datasets}
  \label{photostydataset}
\end{figure}

In this section, we provide a detailed introduction to the style dataset we adopted. As shown in Figure \ref{photostydataset} and Table \ref{tab:stydataset}.

\begin{table}[h]
    \centering
    \caption{style dataset}
    \label{tab:stydataset}
    \begin{tabular}{l|cc}
        \toprule
        \textbf{Photo Index} & \textbf{Subject Prompt} & \textbf{Style Prompt} \\
        \midrule
        \textbf{0101} & A bay & watercolor painting \\
        \textbf{0102} & A house & watercolor painting \\
        \textbf{0103} & A cat & watercolor painting   \\
        \textbf{0104} & Flowers & watercolor painting   \\
        \textbf{0107} & A portrait of a person & oil painting   \\
        \textbf{0301} & Slice of watermelon and clouds in the background & 3d rendering  \\
        \textbf{0304} & A woman & 3d rendering  \\
        \textbf{0305} & A bear & kid crayon drawing\\
        \textbf{0307} & A flower & melting golden 3d rendering   \\
        \textbf{0308} & A Viking face with beard & wooden sculpture\\
        \bottomrule
    \end{tabular}
\end{table}

\subsection{Subject datasets}
\begin{figure}[h]
  \centering
  \includegraphics[width=1\linewidth]{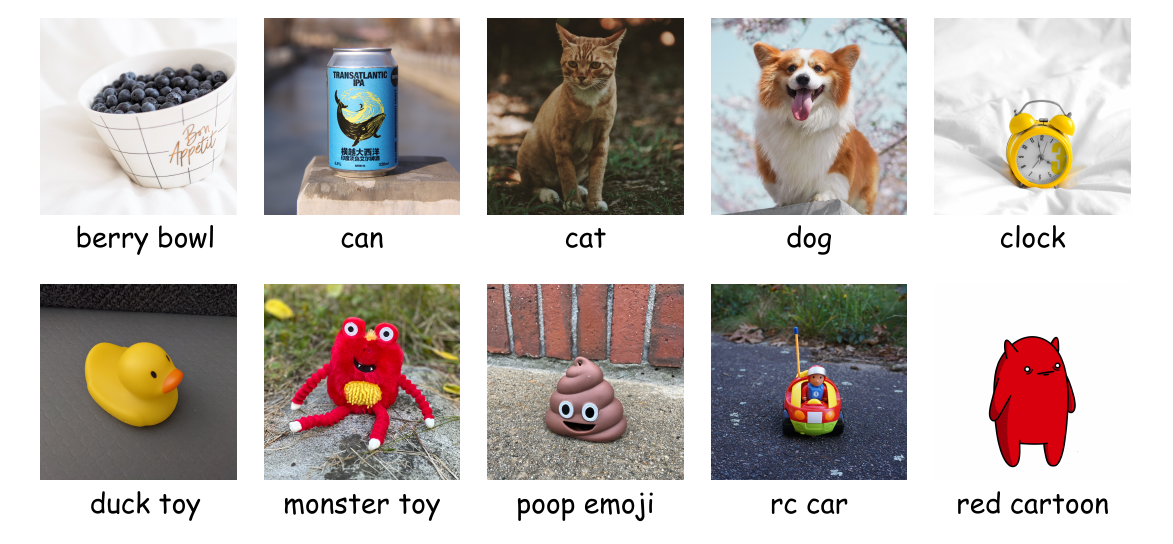}
  \caption{subject datasets}
  \label{photosubdataset}
\end{figure}

In this section, we provide a detailed introduction to the subject dataset we adopted. As shown in Figure \ref{photosubdataset}.

\section{Detailed experimental results}
\begin{figure}[h]
  \centering
  \includegraphics[width=1\linewidth]{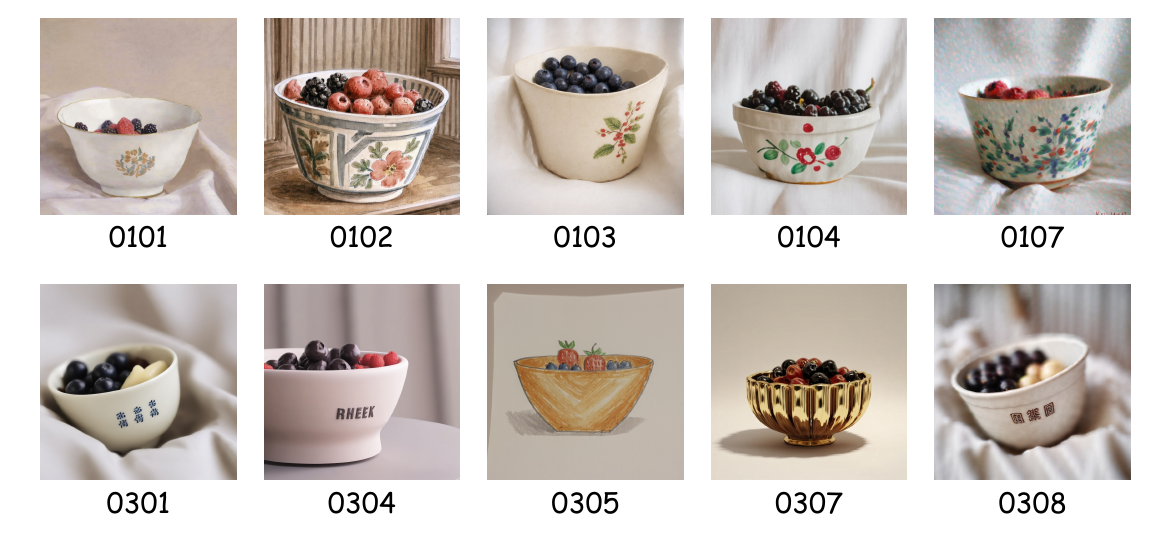}
  \caption{Berry Bowl}
  \label{photobb}
\end{figure}

In this section, we present detailed experimental results. Due to space constraints, we focus on presenting the results for Berry Bowl. As shown in Figure \ref{photobb} and Table \ref{tab:bb}.
\begin{table}[h]
    \centering
    \tabcolsep=0.8cm
    \caption{Berry Bowl}
    \renewcommand\arraystretch{1.3}
    \label{tab:bb}
    \begin{tabular}{l|ccc}
        \toprule
        \textbf{Method} & \textbf{Style Sim} & \textbf{Clip Score} & \textbf{DINO Score}  \\
        \midrule
        \textbf{K-LoRA} & &  &  \\
        \textbf{0101} & 55.35\% & 65.97\%  & 11.45\%  \\
        \textbf{0102} & 60.20\% & 64.99\%  & 17.48\% \\
        \textbf{0103} & 58.65\% & 69.19\%  & 13.29\%  \\
        \textbf{0104} & 68.00\% & 72.88\%  & 33.95\%  \\
        \textbf{0107} & 48.23\% & 58.16\%  & 16.65\%  \\
        \textbf{0301} & 63.72\% & 67.65\%  & 30.76\%  \\
        \textbf{0304} & 54.83\% & 77.05\% & 34.43\%  \\
        \textbf{0305} & 50.35\% & 74.21\% & 13.65\%  \\
        \textbf{0307} & 61.89\% & 76.97\%  & 18.38\%  \\
        \textbf{0308} & 54.83\% & 73.21\% & 4.52\%  \\
        \midrule
        Average & 59.10\% & 73.50\% & 27.50\%  \\
        \midrule
        \textbf{Ours} & &  &  \\
        \textbf{0101} & 51.77\% & 70.24\%  & 9.10\%  \\
        \textbf{0102} & 61.38\% & 60.18\%  & 18.55\% \\
        \textbf{0103} & 57.73\% & 69.67\%  & 17.90\%  \\
        \textbf{0104} & 67.96\% & 72.33\%  & 32.69\%  \\
        \textbf{0107} & 47.32\% & 59.62\%  & 16.80\%  \\
        \textbf{0301} & 65.38\% & 69.48\%  & 31.68\%  \\
        \textbf{0304} & 53.73\% & 77.82\% & 32.72\%  \\
        \textbf{0305} & 54.62\% & 69.42\% & 18.57\%  \\
        \textbf{0307} & 63.16\% & 68.96\%  & 32.69\%  \\
        \textbf{0308} & 54.91\% & 74.82\% & 23.38\%  \\
        \midrule
        average & 62.07\% & 70.00\%  & 32.50\% \\
        \bottomrule
    \end{tabular}
\end{table}

\clearpage
\newpage
\section*{NeurIPS Paper Checklist}

\begin{enumerate}

\item {\bf Claims}
    \item[] Question: Do the main claims made in the abstract and introduction accurately reflect the paper's contributions and scope?
    \item[] Answer: \answerYes{} 
    \item[] Justification: The abstract and introduction accurately encompass the paper's contributions.  A simple yet effective style transfer method that achieves superior performance without training.

\item {\bf Limitations}
    \item[] Question: Does the paper discuss the limitations of the work performed by the authors?
    \item[] Answer: \answerYes{} 
    \item[] Justification: We discuss the limitations of the work in Section~\ref{limitation}.

\item {\bf Theory assumptions and proofs}
    \item[] Question: For each theoretical result, does the paper provide the full set of assumptions and a complete (and correct) proof?
    \item[] Answer: \answerYes{} 
    \item[] Justification: We provide full set of assumptions and a complete proof in Section~\ref{method}.

    \item {\bf Experimental result reproducibility}
    \item[] Question: Does the paper fully disclose all the information needed to reproduce the main experimental results of the paper to the extent that it affects the main claims and/or conclusions of the paper (regardless of whether the code and data are provided or not)?
    \item[] Answer: \answerYes{} 
    \item[] Justification: This article explains the parameter settings and the specific process in Sections~\ref{exp}.

\item {\bf Open access to data and code}
    \item[] Question: Does the paper provide open access to the data and code, with sufficient instructions to faithfully reproduce the main experimental results, as described in supplemental material?
    \item[] Answer: \answerYes{} 
    \item[] Justification: Our code is publicly available at: \url{https://anonymous.4open.science/r/EST-LoRA-F318}.

\item {\bf Experimental setting/details}
    \item[] Question: Does the paper specify all the training and test details (e.g., data splits, hyperparameters, how they were chosen, type of optimizer, etc.) necessary to understand the results?
    \item[] Answer: \answerYes{} 
    \item[] Justification: In the Section~\ref{exp} of this article, all the required training and testing details are clearly defined.

\item {\bf Experiment statistical significance}
    \item[] Question: Does the paper report error bars suitably and correctly defined or other appropriate information about the statistical significance of the experiments?
    \item[] Answer: \answerYes{} 
    \item[] Justification: We generate images using different random seeds and calculate similarity scores, and the results are shown in the Table~\ref{tab:results}.

\item {\bf Experiments compute resources}
    \item[] Question: For each experiment, does the paper provide sufficient information on the computer resources (type of compute workers, memory, time of execution) needed to reproduce the experiments?
    \item[] Answer: \answerYes{} 
    \item[] Justification: In the Section~\ref{exp} of this article, all the required training and testing details are clearly defined.
    
\item {\bf Code of ethics}
    \item[] Question: Does the research conducted in the paper conform, in every respect, with the NeurIPS Code of Ethics \url{https://neurips.cc/public/EthicsGuidelines}?
    \item[] Answer: \answerYes{} 
    \item[] Justification:  All research in this article confirms adherence to the NeurIPS Code of Ethics.

\item {\bf Broader impacts}
    \item[] Question: Does the paper discuss both potential positive societal impacts and negative societal impacts of the work performed?
    \item[] Answer: \answerYes{} 
    \item[] Justification: Our method can be used in generative application scenarios and can mix any style and object without training, which will further accelerate research on image style transfer.
    
\item {\bf Safeguards}
    \item[] Question: Does the paper describe safeguards that have been put in place for responsible release of data or models that have a high risk for misuse (e.g., pretrained language models, image generators, or scraped datasets)?
    \item[] Answer: \answerNA{} 
    \item[] Justification: Our method is fine-tuned by LoRA and is not a pre-trained model, so there is no such risk.

\item {\bf Licenses for existing assets}
    \item[] Question: Are the creators or original owners of assets (e.g., code, data, models), used in the paper, properly credited and are the license and terms of use explicitly mentioned and properly respected?
    \item[] Answer: \answerYes{} 
    \item[] Justification: We make reasonable references to all datasets used.

\item {\bf New assets}
    \item[] Question: Are new assets introduced in the paper well documented and is the documentation provided alongside the assets?
    \item[] Answer: \answerYes{} 
    \item[] Justification: Our code is publicly available at: \url{https://anonymous.4open.science/r/EST-LoRA-F318}.

\item {\bf Crowdsourcing and research with human subjects}
    \item[] Question: For crowdsourcing experiments and research with human subjects, does the paper include the full text of instructions given to participants and screenshots, if applicable, as well as details about compensation (if any)? 
    \item[] Answer: \answerNA{} 
    \item[] Justification: This paper does not involve crowdsourcing nor research with human subjects.

\item {\bf Institutional review board (IRB) approvals or equivalent for research with human subjects}
    \item[] Question: Does the paper describe potential risks incurred by study participants, whether such risks were disclosed to the subjects, and whether Institutional Review Board (IRB) approvals (or an equivalent approval/review based on the requirements of your country or institution) were obtained?
    \item[] Answer: \answerNA{} 
    \item[] Justification: This paper does not involve crowdsourcing nor research with human subjects.

\item {\bf Declaration of LLM usage}
    \item[] Question: Does the paper describe the usage of LLMs if it is an important, original, or non-standard component of the core methods in this research? Note that if the LLM is used only for writing, editing, or formatting purposes and does not impact the core methodology, scientific rigorousness, or originality of the research, declaration is not required.
    \item[] Answer: \answerNA{} 
    \item[] Justification: The core method development in this research does not involve LLMs as any important, original, or non-standard components.

\end{enumerate}

\end{document}